\documentclass[aps,pre,groupedaddress]{revtex4}
\usepackage{amsfonts}
\usepackage{mathrsfs}
\usepackage{graphicx}
\usepackage{subfigure}
\usepackage{amsmath}
\usepackage{amssymb}
\usepackage{amsbsy}
\usepackage{bm}
\usepackage{mathtools}

\begin{document}

\title{Random active path model of deep neural networks with diluted binary synapses}

\author{Haiping Huang}
\email{physhuang@gmail.com}
\affiliation{School of Physics,
Sun Yat-sen University, Guangzhou 510275, People's Republic of China}
\affiliation{Laboratory for Neural Computation and Adaptation, RIKEN Center for Brain Science, Wako-shi, Saitama
351-0198, Japan}
\author{Alireza Goudarzi}
\affiliation{Laboratory for Neural Computation and Adaptation, RIKEN Center for Brain Science, Wako-shi, Saitama
351-0198, Japan}
\date{\today}

\begin{abstract}
Deep learning has become a powerful and popular tool for
a variety of machine learning tasks.
However, it is challenging to understand the mechanism of deep learning from a theoretical perspective. In this work, we propose a random active path model to study collective properties of 
deep neural networks with binary synapses, under the removal perturbation of connections between layers.
In the model, the path from input to output is randomly activated, and the corresponding input unit constrains the weights along the path into the form of a $p$-weight interaction
glass model.
A critical value of the perturbation is observed to
separate a spin glass regime from a paramagnetic regime, with the transition being of the first order. The paramagnetic phase is conjectured to have a poor 
generalization performance. 

\end{abstract}
 \maketitle
\section{Introduction}
Deep learning (e.g., with many layers of neural networks) works very well
in areas from speech recognition, image classification, to drug
discovery, medical image analysis, particle discovery, automatic
game playing, and many others~\cite{DL-2016}. This is due to the available
large dataset for training and efficient hardware design such as GPU to accelerate training, rather
than breakthrough in theoretical foundations. Theoretical efforts are recently devoted to address partially the origin of
these impressive performances of deep networks, relying on simple assumptions~\cite{Lecun-2014,Chau-2015,Kawa-2016,Sou-2016,CNN-2016,Poggio-2018,NIPS-2016}.

One of promising perspectives looking at deep learning is the concept of redundancy, i.e., the deep architecture is robust, to some extent, in the absence of a fraction of connections between layers. In other words, the generalization ability measured by
the output error on test (unseen) dataset does not significantly change until a sufficient amount of connections are removed. Indeed,
it was recently revealed that deep convolutional neural networks mimicking the ventral visual pathway are robust to a number of weight perturbations in the higher convolutional layers~\cite{cnn-2017}.
We also observed that a multi-layer perceptron network classifying handwritten digits has such redundancy behavior (Supplemental Material).
This suggests that the redundancy property may be a general principle of deep computation. In terms of synaptic activities, this kind of
redundancy is expected to appear from interactions between synapses, in a supervised learning system. 

From a theoretical perspective, an interesting question arises, i.e., how the statistical property of a deep neural network changes with 
respect to the removal perturbation of synapses between adjacent layers. We address this question by proposing a simple random active path (RAP) model, in which we construct randomly and independently
each active path from the input at the bottom layer to the output at the top layer ($L$ layers in total), and weights along each path are constrained by the corresponding random input of that path. The RAP model is thus formulated as
a $p$-weight interaction model ($p=L-1$ refers to the depth of the deep network). For simplicity, we assume that the path is randomly activated in the sense that units in each layer are randomly activated with a layer-dependent activation probability and the activation is thus not input-dependent.
The activation is
defined in terms of ReLU function commonly used in deep learning~\cite{Nair-2010}. In addition, the activation of a path also depends on the removal perturbation of synapses that can make a
path inactivated. One significant difference from previous $p$-spin glass models~\cite{Gardner-1985,Kirk-1987,Franz-2001,Montanari-2003} is that
the degree (the number of paths one weight is involved in) distribution of weights in the model depends on both the layer-dependent activation probability and the perturbation level of 
deep neural networks, such that there are multiple peaks in the distribution, and the degree can vary in a wide range of values. 

The RAP model can be analytically studied by mean-field methods, revealing a critical value of the perturbation separating a paramagnetic phase from 
a spin glass phase. The paramagnetic phase is conjectured to be related to the non-robust regime of deep neural networks under weight perturbations. This result may also explain the performance of the dropconnect algorithm~\cite{dropcon}, 
where a finite fraction of connections is stochastically turned off during training, when different fractions are used.

The paper is structured as follows.  We first introduce the RAP model to study
collective properties of deep neural networks with binary synapses, by varying the perturbation level. Theoretical predictions match very well Monte-Carlo simulation results.
We give concluding remarks and future perspectives in the final section.

\begin{figure}
\centering
    \includegraphics[bb=23 391 584 693,scale=0.4]{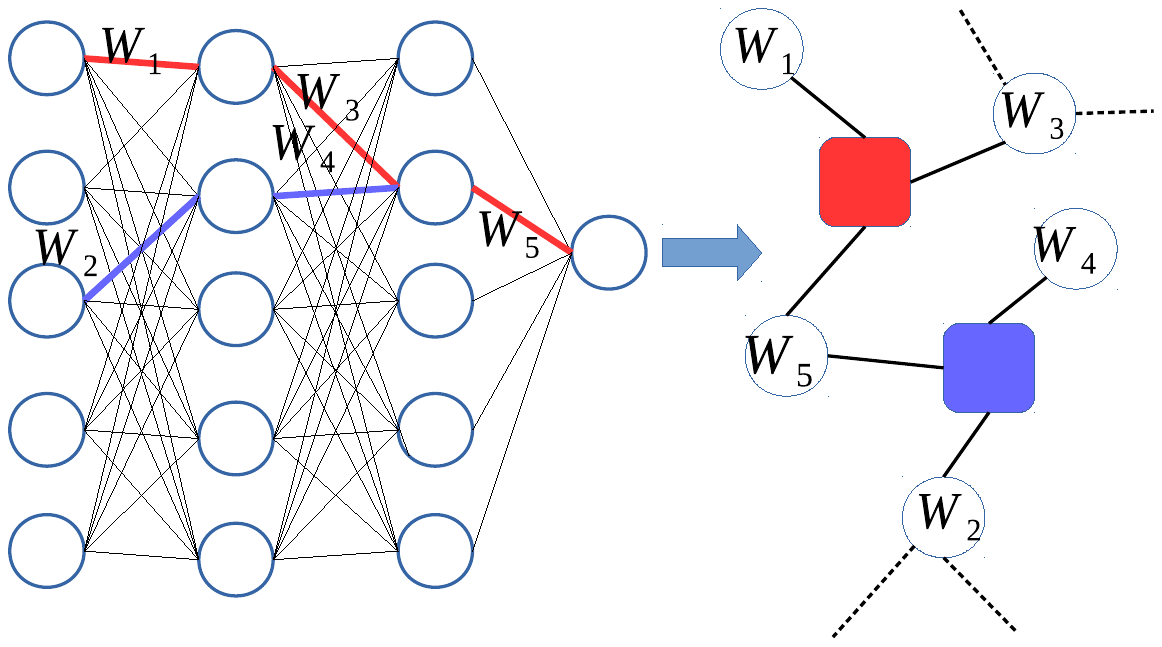}
  \caption{ (Color online) Schematic illustration of a random construction of the RAP model. Left panel: a $4$-layer deep network has an equal width in each layer (except the last layer). Two active paths are selected and share a common weight ($W_5$). Circle nodes indicate ReLU units.
  Right panel: two constraints denoted by square nodes are present in the factor graph of the Hamiltonian (Eq.~(\ref{rap})) and correspond to the two active paths in the left panel. Circle nodes denote the weights on each path. The weights may be connected to other
  constraints (indicated by dash lines).
  }\label{rapmod}
\end{figure}

\begin{figure}
\centering
    \includegraphics[bb=0 0 312 244,scale=0.8]{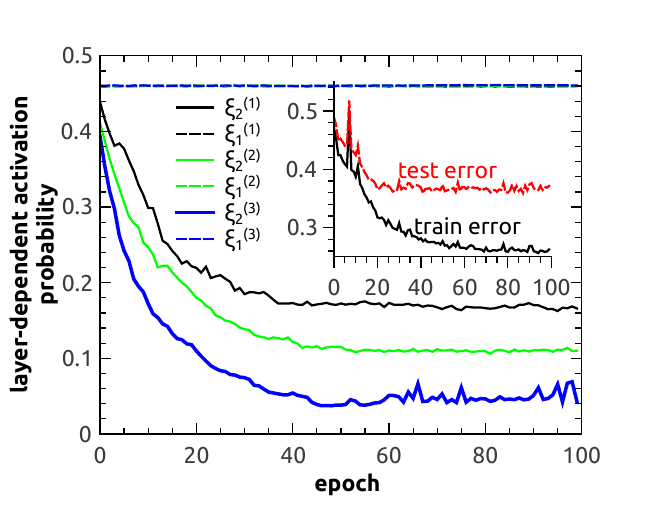}
  \caption{ (Color online) Layer-dependent activation probability as a function of training epochs. Three independent training trials (indicated by superscripts of $\xi_{1,2}$) are shown. The inset shows one example of 
  how training and test errors change with epochs. The deep network with the same architecture as in Fig.~\ref{rapmod} ($n_0=n_1=n_2=100$) is trained using the absolute loss, in which the binary target
  is generated from a teacher network. We used Adam~\cite{Adam} during stochastic gradient descent of the loss function. $5000$ examples are used for training, while the other $1000$ examples are used for testing.
  Results do not qualitatively change when different optimization methods (e.g., reinforced stochastic gradient descent~\cite{Huang-2017b}) are used.
  }\label{actprob}
\end{figure}

\section{A random active path model}
\label{model}
In this section, we propose a random active path model 
to \textit{qualitatively} understand the collective behavior of deep neural networks with diluted binary synapses, i.e., to address how the statistical property of a deep neural
network changes with respect to the removal perturbation of synapses between adjacent layers. We consider a deep network model with $L=4$ layers of fully-connected feedforward architecture. 
Each layer has $n_l(l=0,\ldots,L-1)$ units (so-called width of that layer). The deep network architecture is specified by $n_{0}$-$n_{1}$-$n_{2}$-$n_{3}$. The input is defined as an $n_0$-dimensional vector $\mathbf{v}$, 
and the weight matrix $\mathbf{W}^l$ with binary ($\pm1$) entries specifies connections between layer $l$ and layer $l-1$. 
A bias can be incorporated into the weight matrix by assuming an additional constant input unit. 
The output at the final layer is expressed as:
\begin{equation}\label{output}
    \mathbf{y}=f_{L-1}\left(\mathbf{W}^{L-1}f_{L-2}(\mathbf{W}^{L-2}\cdots f_1(\mathbf{W^1}\mathbf{v}))\right),
\end{equation}
where $f_l(\cdot)$ is an element-wise ReLU function for units at layer $l$, defined as $f_l(u_i)={\rm max}(0,u_i)$ where $u_i$ is the weighted-sum input to the unit $i$ at layer $l$. 
For simplicity, we assume that all layers except the top one use ReLU transfer function, and the top layer has only one output unit that
uses an identity map~\cite{Kawa-2016}.

An active path refers to the path from one input unit to one output unit with the property that each connection on the path
is present (each synapse is deleted with a dilution probability $p_l$, where $l$ specifies the weight population between two consecutive layers ($l$ and $(l-1)$-layer)) and all units along the path are activated because of ReLU activation.
It follows that the network output defined in Eq.~(\ref{output}) can be re-expressed as~\cite{Lecun-2014,Chau-2015}
\begin{equation}\label{output2}
    y=\sum_{a=1}^{\Psi}v^{a}\prod_{k=1}^{L-1}W^{k}_{a},
\end{equation}
where $\Psi$ denotes the total number of active paths converged to the output unit, $v^{a}$ denotes the input specified to the $a$-th path for the output, and $W^{k}_{a}\in\{\pm1\}$ denotes 
the entry of $\mathbf{W}^{k}$ that is used in the $a$-th path for the output.

A random construction of active paths is shown in Fig.~\ref{rapmod}, where only active paths are used to
construct the Hamiltonian of the model composed of many constraints (represented by square nodes), and each of these constraints (e.g., $a$) is given by the input $v^a$ specified to the corresponding active path. In this construction process,
we assume that activation of paths is independent of input~\cite{Lecun-2014}, realized by the units of each hidden layer activated independently with an activation probability, say $\xi_l$ that is layer-dependent and can be 
empirically estimated from deep neural network trainings (Fig.~\ref{actprob}). The Hamiltonian of
the RAP model can thus be written as
\begin{equation}\label{rap}
   \mathcal{H}(\mathbf{W})=-\sum_{a=1}^{n^3}A_av^a\prod_{i\in\partial a}W_i,
\end{equation}
where $n$ denotes the layer width (the same for all layers except the last one), $A_a$ is a binary value indicating the activation ($A_a=1$) or silent ($A_a=0$), and $\partial a$ denotes the set of the weights involved in the active path $a$. 
According to the random construction, the probability of a constructed active path is defined by $P(A_a=1)=\prod_l\xi_l(1-p_l)$, where $\xi_l$ and $p_l$ are aforementioned activation and dilution probabilities, respectively.
Note that the weight configuration $\mathbf{W}$ in the model is a subset of 
$\{\mathbf{W}^l\}$ in Eq.~(\ref{output}) unless all $n^3$ paths of the deep network become active. Here, we assume that $v^a$ are randomly selected from components of a random input vector $\mathbf{v}$ whose components follow independently a binary distribution
($P(v_i=\pm1)=0.5$). During the random construction, some constraints may share the same value of $v^a$, i.e., the corresponding paths share a common input unit. 

To build a naive relationship between the Hamiltonian we studied here and the loss function (e.g., absolute loss) used in training (Fig.~\ref{actprob}), we assume that the true label is denoted by
$Y_t=\pm\Lambda$ where $\Lambda>0$ is a predefined maximal output. 
We then define $\mathcal{H}=-{\rm sgn}(Y_t)y$. It is easy to verify that the learning process minimizing the absolute loss ($\mathcal{C}=|Y_t-y|$) between the target and the actual output $y$ is equivalent to
finding the minimal value of $\mathcal{H}$ in the RAP model. If we assume $Y_t$ is random, ${\rm sgn}(Y_t)$ can be absorbed into the input and the model is statistically invariant.
The random hinge loss can also be similarly analyzed.

Although it is challenging to prove the assumptions we made above reasonable in practical deep networks, the above $p$-weight interaction model still
provides us a nice starting point to qualitatively understand complicated properties of deep neural networks. 
We will show interesting (non-trivial) properties of this simple model below. 

\begin{figure}
\centering
\subfigure[]{
   \includegraphics[bb=0 0 337 244,width=0.45\textwidth]{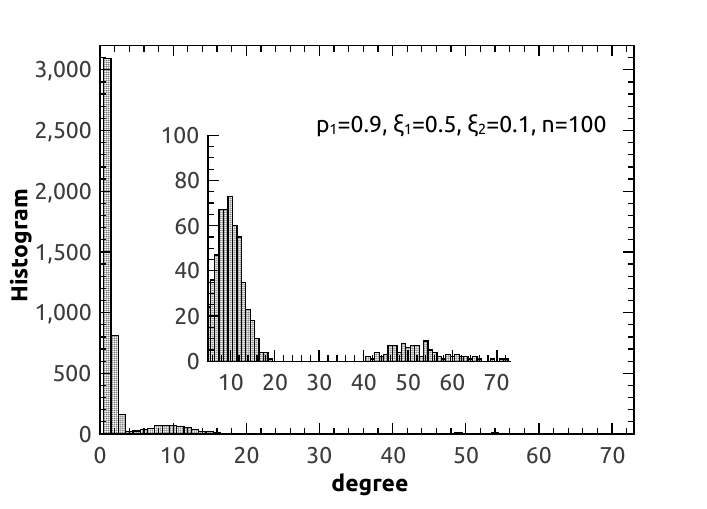}}
      \hspace{0.01cm}
  \subfigure[]{\includegraphics[bb=0 0 325 244,width=0.45\textwidth]{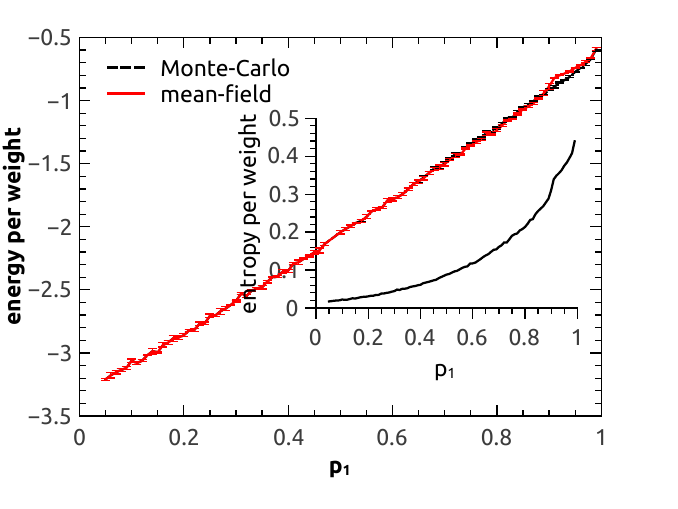}}\\
   \vspace{0.005cm}
      \subfigure[]{\includegraphics[bb=0 0 325 244,width=.45\textwidth]{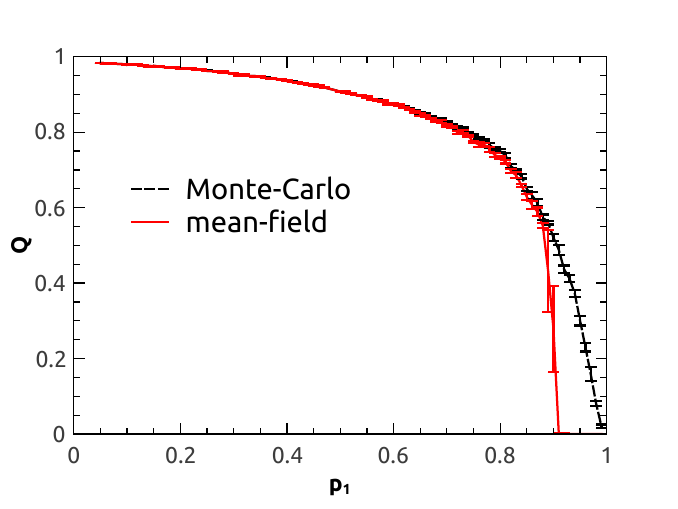}}\\
  \caption{(Color online) Statistical properties of the RAP model. We consider the case of $\xi_1=0.5$, $\xi_2=0.1$, and $\beta=1$. The result is averaged over five random samples of the model with the maximal number of weights equal to $2n^2+n=20100$. Monte-Carlo simulation results are compared. 
  (a) The degree distribution of weights in a random instance.
  The inset is an enlarged view. The qualitative behavior does not change when another instance is generated.
  (b) Energy per weight as a function of the dilution probability $p_1$. The entropy is shown in the inset. 
  (c) Spin glass order parameter ($Q=\frac{1}{N_w}\sum_im_i^2$, where $N_w$ is the total number of weights in the model) versus the dilution probability $p_1$. The magnetization $m_i$ can be estimated by either mean-field mehtods or Monte-Carlo sampling.
     }\label{rapE}
 \end{figure}

We first evaluate the empirical values of the layer-dependent activation probability ($\xi_l$) from training a deep neural network with the same structure as our model (Fig.~\ref{rapmod}). The only difference is that,
the binary target at the top layer is generated by a teacher network with the same structure fed with the corresponding input. The result is shown in Fig.~\ref{actprob}, which implies that the first hidden layer has a higher activation probability (nearly independent of trials) than
the subsequent layer. The activation probability in the subsequent layer varies in a relatively large range, suggesting a sparse distributed representation for efficient computation in the deep ReLU network~\cite{sparse-2011}.
Therefore, in our model analysis, we fix $\xi_1=0.5$, and $\xi_2=0.05,0.1,0.15$. In fact, the value of $\xi_1=0.5$ can be derived by assuming a normal distribution for the weighted-sum $u_i$.
The model defined in Eq.~(\ref{rap}) can be expressed as a factor graph like the one shown
in Fig.~\ref{rapmod}. The degree of a weight is thereby defined as the number of constraints the weight is involved in. The degree distribution of weights in a typical example is shown 
in Fig.~\ref{rapE} (a), which suggests that there exists three peaks corresponding to weights drawn from three different weight populations. The well-structured degree distribution is a feature of the RAP model constructed to
mimic the behavior of a deep neural network.

The RAP model (Eq.~(\ref{rap})) is thus different from previous $p$-spin glass models~\cite{Gardner-1985,Kirk-1987,Franz-2001,Montanari-2003} in three aspects. First, the model 
is interpreted in terms of active-path decomposition of a deep network's output that is directly connected to training/test errors. 
In particular, the binary activation indicator $A_a$ is controlled by two deep-network-dependent parameters---the dilution probability and layer-dependent activation probability of units.
Second, when an active path is randomly constructed, the quenched disorder comes from the corresponding input unit, thereby being physically explained. Lastly, the constructed model has a distinct three-peak degree distribution in the corresponding graphical representation,
and each peak reflects the feature of a different layer.

 \begin{figure}
\centering
\subfigure[]{
   \includegraphics[bb=0 0 325 244,width=0.45\textwidth]{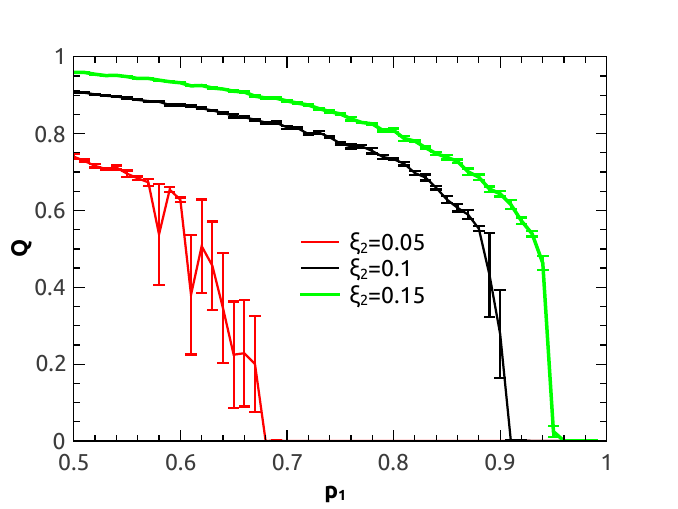}}
      \hspace{0.01cm}
  \subfigure[]{\includegraphics[bb=0 0 325 244,width=0.45\textwidth]{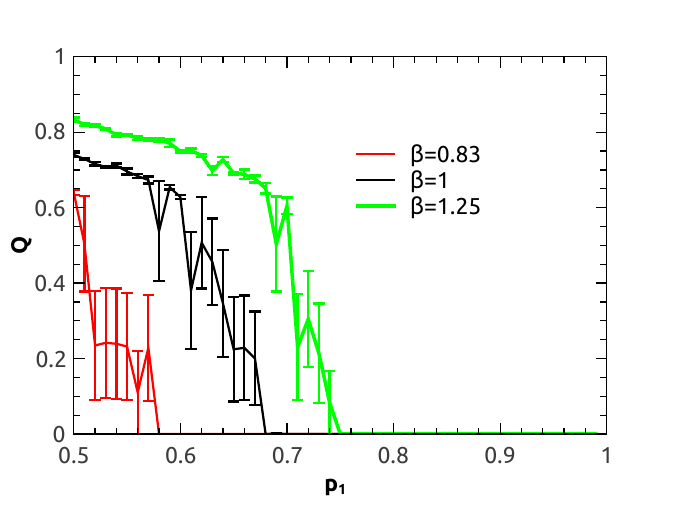}}\\
  \caption{(Color online) Spin glass order parameter versus the dilution probability, when the activation probability $\xi_2$ (a) and the inverse-temperature (b) are varied. In (a), the temperature is fixed to one, while $\xi_2$ is fixed to 0.05 in (b). Other parameters are the same as in Fig.~\ref{rapE}.
  The results are computed by the mean-field method.}
  \label{rapSG}
 \end{figure}
The mean-field model defined in Eq.~(\ref{rap}) 
can be solved by the cavity method~\cite{cavity-2001,Huang-2016a}. The technical details are summarized in Appendix~\ref{CMrap}. We consider the case of varying $p_1$ with zero $p_2$ and $p_3$. Because of layer-dependent activation probability,
the quantitative behavior depends on which weight population is diluted.
As shown in Fig.~\ref{rapE} (b) and (c), given the inverse temperature $\beta=1$, 
the RAP model has a paramagnetic phase when $p_1$ is larger than $0.9$, and the energy keeps decreasing as $p_1$ decreases. The theoretical prediction of the cavity method is in agreement with the simple Monte-Carlo simulations of the model (simulation details are given in Appendix~\ref{MCrap}),
despite deviations around the transition point. The deviations may be caused by the nature of the transition and sampling accuracy of the quenched Monte-Carlo procedure. First, the transition is of the first-order (Fig.~\ref{rapE} (c)), because there exists metastability of states (paramagnetic state versus spin glass state in which weights prefer to
polarize towards a particular direction, but a spatial average over all magnetizations vanishes). Therefore, even in paramagnetic phase, depending on the details of the evolution, the quenched Monte-Carlo dynamics has a certain probability to be attracted by the spin glass state, which seems to have a lower energy than the paramagnetic state yet a weak contribution to the equilibrium measure in the paramagnetic phase (in terms of free energy taking into account the entropy contribution).
Second, the accuracy of the Monte-Carlo sampling is affected by the number of collected configurations and relaxation steps, and how abruptly the energy landscape is changed.
The effect is expected to be stronger around the transition point than the regime where one phase already overwhelms the other, which is reflected by the large error bars in Fig.~\ref{rapE} (c). 

As shown in the inset of Fig.~\ref{rapE} (b), the entropy increases when more and more weights are pruned. In this case of large $p_1$,
the deep network system becomes less constrained with a large number of candidate weight configurations, but the energy is relatively high.
The entropy curve also displays a slight jump (about $0.03$ when $p_1$ increases from $0.9$ to $0.91$), which is a characteristic of the first-order phase transition.
Overall, the equilibrium properties of the RAP model demonstrate that reducing the dilution probability will trigger a phase transition towards a spin glass state, and the paramagnetic phase located at a high dilution probability is clearly not favored,
because it does not support the generalization ability of the network (one does not expect a randomly polarized (with equal probability) yet equilibrium configuration of weights yields
a good generalization), and thus may connect to
the high test errors when a sufficiently large fraction of connections are removed (Supplemental Material).

Finally, we explore the effects of different inverse-temperatures and intermediate-layer activation probabilities on the phase transition in Fig.~\ref{rapSG}. When the activation probability decreases, the critical dilution probability decreases as well.
This implies that the regime of the paramagnetic phase expands, impairing the generalization performance of the network when an intermediate number of weights are deleted.
On the other hand, the inverse-temperature is a measure of the stochastic noise level in learning process. A high temperature also expands the paramagnetic phase, thereby impairing the network's robustness.

\begin{figure}
\centering
\subfigure[]{
   \includegraphics[bb=0 0 398 333,width=0.45\textwidth]{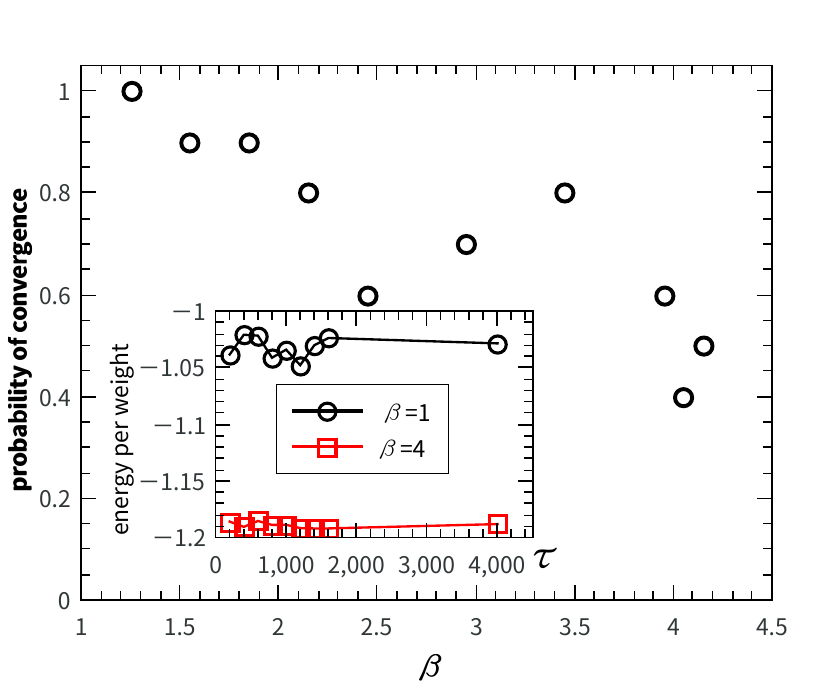}}
      \hspace{0.01cm}
  \subfigure[]{\includegraphics[bb=0 0 410 334,width=0.45\textwidth]{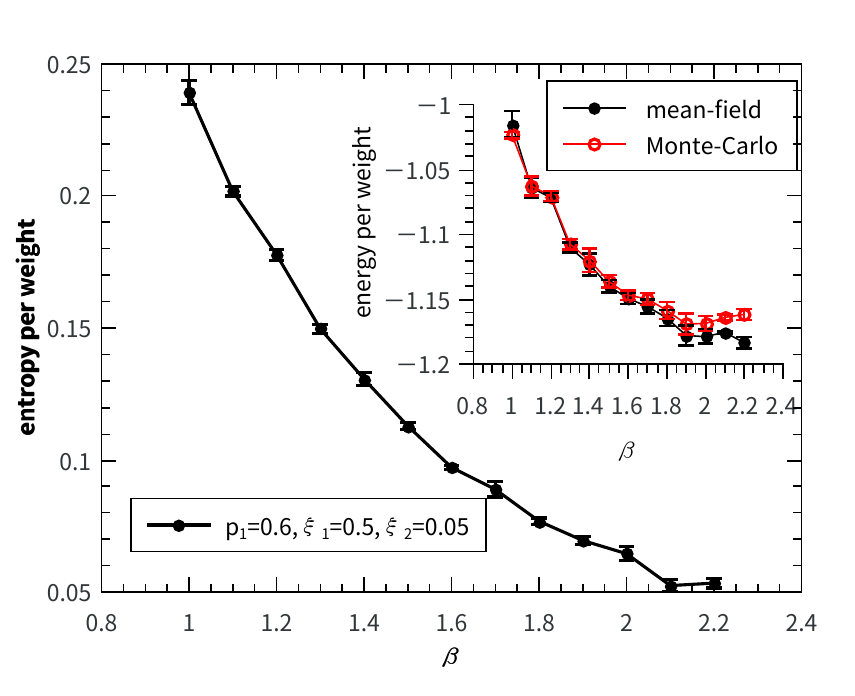}}\\
  \caption{(Color online) (a) Stability of the mean-field iteration and the energy levels reached by the simulated annealing where cooling rates ($\tau$) are tuned (see Appendix~\ref{MCrap}).
  The stability is measured by the fraction of the iteration convergence among ten trials.
  $p_1=0.6$. $\xi_2=0.05$. Other parameters are the same as in Fig.~\ref{rapE}.
  (b) Entropy per weight changes with inverse-temperatures. The inset shows the energy per weight. Other parameters are the same as in Fig.~\ref{rapE}.}
  \label{rapstab}
 \end{figure}

We further investigate the effects of inverse-temperatures on the validity of our results. With decreasing the temperature, the iteration of mean-field equations (Appendix~\ref{CMrap}) will lose its stability (Fig.~\ref{rapstab} (a)), for example, exhibiting an oscillatory behavior.
In this case, the weight space may develop complex structures, i.e., the low-energy configurations form exponentially many clusters, and they may further condense onto a finite number of states, especially when the temperature is further lowered down. Then we should adopt the replica symmetry breaking assumption
to study this complex situation (Appendix~\ref{CMrap}). However, within the temperature range we explore in Fig.~\ref{rapE} and Fig.~\ref{rapSG}, the iteration is stable, leading to a converged value for the order parameter. Moreover, the entropy does not become negative (Fig.~\ref{rapstab} (b)).
The energy level predicted by the mean-field method coincides well with that reached by the Monte-Carlo dynamics, although at a slightly low temperature, the mean-field method, compared with the Monte-Carlo dynamics,
can reach a lower energy level (the inset of Fig.~\ref{rapstab} (b)). We also perform simulated annealing experiments on the constructed model to study the effects of 
cooling rates on the final reached energy levels (technical details are given in Appendix~\ref{MCrap}). Interestingly, when the cooling becomes slower, the energy level the dynamics reaches does not show 
a significant change (Fig.~\ref{rapstab} (a)), indicating that a dynamical ergodicity-breaking transition is not apparent.

In this work, we only consider $4$-layer feedforward neural networks, which correspond to a three-body interaction spin model. It is interesting to study the infinite depth limit, because in a standard
$p$-spin interaction model where the coupling between spins follows a Gaussian distribution, and all possible coupling among $p$ spins are considered, Derrida showed that in the large-$p$ limit,
the model behaves like a random energy model, in which the joint energy level distribution becomes factorized, and a frozen phase can be identified when the temperature is lowered down~\cite{REM-1981}. In our current
context, according to the techniques used in Refs.~\cite{REM-1981,Domi-1987}, the joint distribution of $\mathcal{N}=2^{N_{w}}$ energy levels can be expressed as
\begin{equation}\label{remnn}
 P(E_1,E_2,\ldots,E_{\mathcal{N}})=\int_{-\infty}^{+\infty}\prod_{a=1}^{\mathcal{N}}\frac{d\hat{E}_a}{2\pi}\exp\left(\sum_a{\rm i}\hat{E}_aE_a+n^p\prod_l\xi_l(1-p_l)\Bigl(\prod_a\cosh({\rm i}\hat{E}_a)-1\Bigr)\right),
\end{equation}
where $n^p\prod_l\xi_l(1-p_l)$ is exactly the number of active paths in the model, and the number is expected to be finite when $p$ tends to be infinite. It follows that the single energy level follows a Gaussian distribution with
a fluctuation of the order of $[n^p\prod_l\xi_l(1-p_l)]^{1/2}$ around zero, to leading order. To derive Eq.~(\ref{remnn}), we assume that the overlap $q$ between two configurations is smaller than one in the 
magnitude, therefore $q^p$ is a negligible term. We also use the gauge transformation~\cite{Mezard-1984}. It is not 
immediate to conclude that the energy levels are independent random variables from Eq.~(\ref{remnn}). Studying the neural network in the infinite depth limit is very interesting.
In practice, training a very deep network is not easy due to the gradient vanishing problem. However, in the infinite depth limit, the energy levels may be organized into a non-trivial structure (e.g., a hierarchical organization like the one in glassy systems). The detailed exploration of this limit is beyond
the current scope, and we leave that for future works.

\section{Discussion}
\label{Disc}
The deep architecture is expected to be robust in
the absence of a small fraction of connections between layers. In practice, not all connections (synapses) are used for transforming the original input, given that the number of parameters in modern deep learning architectures is typically
much larger than the amount of training data. Given the redundancy property, the deep network has the chance of
repairing and modification of subnetworks, without affecting the overall function. It was recently revealed that deep convolutional neural networks mimicking the ventral visual pathway are robust to a number of weight perturbations in the higher convolutional layers~\cite{cnn-2017}.
This suggests that the redundancy property may be a general principle of deep computation.

Motivated by these empirical observations, we propose the RAP model to understand how the collective properties of deep neural networks change with the removal perturbation.
Reducing the number of active paths drives the network to a paramagnetic phase where the weights behave like a noise with zero mean. In the opposite direction, the network 
enters a spin glass regime through a first-order phase transition. In the spin glass phase, the weights polarize to their own particular directions, rather than fluctuating with zero mean.
This regime is meaningful in the sense that the network learns the feature of the input-output relationship hidden in the training data, with a configuration of particularly-polarized weights, which may connect to the robustness property of deep networks.
In contrast, the paramagnetic phase does not have such properties, since weights behave like a random guess (taking both directions with equal probability). In other words, the redundancy is completely removed. In addition, the nature of the transition implies that 
around the transition, the paramagnetic phase is still stable and competes with the spin glass phase, which predicts that the learning should be difficult when the hyper-parameter such as dilution probability is chosen to be around the 
transition point.

Our model may connect to the dropconnect technique~\cite{dropcon}, one popular regularization technique used during training deep networks. The dropconnect was proposed to stochastically turn off a finite fraction of connections during training, providing a sampling of model ensemble 
in the course of the training, thereby reducing over-fitting effects. When the dropconnect probability (the fraction of connections kept during training) is very small, the performance gets worse rapidly as the probability further decreases~\cite{dropcon},
which is consistent with the paramagnetic phase as explained above. This suggests that one should choose
a relatively large dropconnect probability to implement a sampling of good model ensemble in terms of its robustness in the learning performance.

The RAP model still needs to be improved by considering how activation of paths relies on the (structured) input and its specific label, together with
conditional dependence of units' activity at one layer on their previous layer. Thus this work encourages further refined models that can have
quantitative predictions of the collective behavior observed in practical applications of deep learning. On the other hand, when the learning noise is weak, which corresponds
to using a low temperature to explore the model's properties, the model may exhibit rich equilibrium/dynamical properties, for which replica symmetry breaking calculations or glassy dynamics analysis may provide
theoretical insights. These insights may also clarify how the robustness property of a deep network under training is affected. We leave this for future works.

\appendix
\begin{widetext}
\section{Mean field method to solve the RAP model}
\label{CMrap}
We present self-consistent mean-field equations in this appendix to analyze the statistical properties of the RAP model (Eq.~(\ref{rap})). These equations can be derived using the 
standard cavity method~\cite{MM-2009,Huang-2016a}. First, the weight configuration $\mathbf{W}$ follows
the Boltzmann distribution $P(\mathbf{W})\propto e^{-\beta\mathcal{H}(\mathbf{W})}$, where the normalization constant defined by $Z$ is the partition function of the model, and the inverse-temperature $\beta$ determines 
the energy level one is interested in. We first calculate the partition function of a new system obtained by
adding a weight and its associated $C$ constraints (the degree of this weight is $C$), as given by
\begin{equation}\label{Z_Node}
    \begin{split}
    Z^{new}&=\sum_{W_{i}}\sum_{\mathbf{W}}\exp\bigl({\sum_{a=1}^{M}\beta v^{a}\prod_{k\in\partial a}W_{k}+
    \beta\sum_{b=1}^{C}v^{b}W_{i}\prod_{j\in\partial b\backslash i}W_{j}}\bigr)\\
    &=Z^{old}\sum_{W_{i}}\prod_{b}\sum_{\{W_{j}\}:j\in\partial b\backslash i}\prod_{j\in \partial b\backslash i}
    \biggl[\frac{1+W_jm_{j\rightarrow b}}{2}\biggr]\cdot e^{\beta v^{b}W_{i}\prod_{j\in \partial b\backslash
    i}W_{j}}\\
    &=Z^{old}\biggl[\prod_{b}\bigl[\cosh\beta v^{b}\bigl(1+\tanh\beta v^{b}\prod_{j\in\partial b\backslash i}m_{j\rightarrow
    b}\bigr)\bigr]+\prod_{b}\bigl[\cosh\beta v^{b}\bigl(1-\tanh\beta
v^{b}\prod_{j\in\partial b\backslash i}m_{j\rightarrow
b}\bigr)\bigr]\biggr],
    \end{split}
\end{equation}
where $M$ denotes the total number of active paths (constraints) in the model, $W_{i}$ is the newly added weight,
$Z^{old}=\sum_{\mathbf{W}}\exp({\sum_{a=1}^{M}\beta v^{a}\prod_{i\in
\partial a}W_{i}})$ is the partition function of the old system,
 $m_{j\rightarrow b}$ is the cavity magnetization of weight $j$ in the absence of constraint $b$, and $j\in\partial
b\backslash i$ denotes the set of weights involved in the constraint $b$
but $i$ is excluded from this set. To derive the second equality in
Eq.~(\ref{Z_Node}), we use the assumption that when a constraint is removed, its neighboring weights become independent, 
which is asymptotically correct when the factor
graph is locally tree-like. Upon defining the conjugate magnetization
$\hat{m}_{b\rightarrow i}\equiv\tanh\beta v^{b}\prod_{j\in\partial b\backslash i}m_{j\rightarrow b}$, one gets the free energy shift:
\begin{equation}\label{deltaFi}
    -\beta\Delta F_{i}=\ln\frac{Z^{new}}{Z^{old}}=\ln\biggl[\prod_{b\in\partial i}\bigl[\cosh\beta v^{b}(1+\hat{m}_{b\rightarrow i})
    \bigr]+\prod_{b\in\partial i}\bigl[\cosh\beta v^{b}(1-\hat{m}_{b\rightarrow i})\bigr]\biggr].
\end{equation}

Then, we add a constraint (e.g., $a$) to form a new system, and calculate the new partition function as
\begin{equation}\label{Z_Func}
    \begin{split}
    Z^{new}&=\sum_{\mathbf{W}}\exp\bigl({\beta\sum_{b=1}^{M}v^{b}\prod_{k\in \partial
    b}W_{k}+\beta v^{a}\prod_{i\in\partial a}W_{i}}\bigr)\\
    &=\sum_{\mathbf{W}}e^{\beta\sum_{b=1}^{M}v^{b}\prod_{k\in \partial
    b}W_{k}}\sum_{\mathbf{W}}\frac{e^{\beta\sum_{b=1}^{M}v^{b}\prod_{k\in \partial
    b}W_{k}}}{\sum_{\mathbf{W}}e^{\beta\sum_{b=1}^{M}v^{b}\prod_{k\in \partial
    b}W_{k}}}e^{\beta v^{a}\prod_{i\in \partial a}W_{i}}\\
    &=Z^{old}\sum_{\mathbf{W}}P(\mathbf{W})e^{\beta v^{a}\prod_{i\in \partial
    a}W_{i}}\\
    &=Z^{old}\sum_{\{W_{i}\}:i\in\partial a}\prod_{i\in\partial
    a}\biggl[\frac{1+m_{i\rightarrow a}W_{i}}{2}\biggr]e^{\beta v^{a}\prod_{i\in \partial
    a}W_{i}}\\
    &=Z^{old}\cdot\cosh\beta v^{a}\bigl(1+\tanh\beta v^{a}\prod_{i\in\partial a}m_{i\rightarrow a}\bigr).
    \end{split}
\end{equation}
The corresponding free energy shift is thus $-\beta\Delta
F_{a}=\ln\biggl[\cosh\beta v^{a}\bigl(1+\tanh\beta
v^{a}\prod_{i\in\partial a}m_{i\rightarrow a}\bigr)\biggr]$.
Finally, the Bethe free energy scaled by $\beta$ can be constructed as follows~\cite{cavity-2001}:
\begin{equation}\label{freeE}
F\equiv-\ln Z=\sum_{i}\Delta F_i-\sum_{a}(|\partial a|-1)\Delta F_a.
\end{equation}

Following a variational principle of the free energy, one can derive the recursive equation of $\{m_{i\rightarrow a},\hat{m}_{b\rightarrow i}\}$ as follows:
\begin{subequations}\label{bp}
\begin{align}
m_{i\rightarrow a}&=\tanh\left(\sum_{b\in\partial i\backslash a}\tanh^{-1}\hat{m}_{b\rightarrow i}\right),\\
\hat{m}_{b\rightarrow i}&=\tanh \beta v^b\prod_{j\in\partial b\backslash i}m_{j\rightarrow b},
\end{align}
\end{subequations}
where $\partial b\backslash i$ denotes the member of interaction $b$ except $i$, and $\partial i\backslash a$ denotes the interaction set
$i$ is involved in with $a$ removed. $m_{i\rightarrow a}$ is interpreted as the message passing from the weight $i$ to the interaction $a$ it
participates in, while $\hat{m}_{b\rightarrow i}$ is interpreted as the message passing from the interaction $b$ to its member $i$. We call this iteration equation
the belief propagation, which serves as the message passing algorithm whose fixed point corresponds to the stationary point of the Bethe free
energy, from the variational principle~\cite{MM-2009}.

With the free energy, one can estimate both
the entropy and the energy of the model. The energy is given by:
\begin{subequations}\label{Energ}
\begin{align}
E&=-\sum_{i}\Delta E_i+\sum_{a}(|\partial a|-1)\Delta E_a,\\
\Delta E_i&=\frac{\sum_{x=\pm1}\mathcal{G}_i(x)}{\sum_{x=\pm1}\mathcal{H}_i(x)},\\
\Delta E_a&=v^a\frac{\tanh \beta v^a+\prod_{i\in\partial a}m_{i\rightarrow a}}{1+\tanh \beta v^a\prod_{i\in\partial a}m_{i\rightarrow a}},\\
\begin{split}
\mathcal{G}_i(x)&=\sum_{b\in\partial i}\Biggl[v^b\sinh \beta v^b(1+x\hat{m}_{b\rightarrow i})+x v^b\cosh\beta v^b(1-\tanh^{2}\beta v^b)\\
&\times\prod_{j\in\partial b\backslash i}m_{j\rightarrow b}
\Biggr]\prod_{a\in\partial i\backslash b}\cosh\beta v^a(1+x\hat{m}_{a\rightarrow i}),
\end{split}\\
\mathcal{H}_i(x)&=\prod_{b\in\partial i}\cosh\beta v^b(1+x\hat{m}_{b\rightarrow i}).
\end{align}
\end{subequations}
The entropy can be obtained by using the standard thermodynamic formula as $S=-F+\beta E$. In the paramagnetic phase, all magnetizations vanish, and analytic expressions for both energy and entropy can be derived.
The energy per weight is written as $-\alpha\tanh\beta$, where $\alpha$ denotes the ratio between the number of constructed active paths and the number of weights involved in these paths,
and the value of $\alpha$ depends on both $p_l$ and $\xi_l$, and thus can be estimated after the random construction of the RAP model. Analogously, the entropy per weight is written as
$\ln2+\alpha(\ln\cosh\beta-\beta\tanh\beta)$.

The above equation is derived under the replica symmetry assumption, i.e., the Gibbs state does not split into an exponential number of states where each of them (indexed by $\gamma$) is described by a free energy
$F^\gamma$. If this assumption is broken, a replica symmetry breaking picture should be introduced. Under this picture, the message passing along each edge (e.g., $i\rightarrow a$) of the factor graph 
turns out to be a probability density $P_{i\rightarrow a}(m_{i\rightarrow a})$, capturing the fluctuations of a replica symmetric message among different states. The replica symmetry breaking case can be in principle analyzed, yet
complicates the analysis, therefore
we would not give a full introduction here, and interested readers can find basics in the classic book~\cite{MM-2009}. We remark here that the replica symmetry solution seems enough to describe the thermodynamic behavior of the
investigated model, within the explored range of parameters, which is supported by the stability of the solution and the non-negative entropy for a discrete system.
\end{widetext}
\section{Monte-Carlo method to simulate the RAP model}
\label{MCrap}
Given the model defined in Eq.~(\ref{rap}), one can explore its energy landscape by using the Monte-Carlo importance sampling method. More precisely, after a random initialization of the weight
configuration, the configuration is updated by the following rule: the transition probability from state $\mathbf{W}$ to $\mathbf{W}'$ with only $W_i$ flipped ($W_i'=-W_i$) is expressed as
$\exp(-2\beta W_iH_i)$ where the effective local field is defined as $H_i=\sum_{a\in\partial i}v^a\prod_{j\in\partial a\backslash i}W_j$. In our Monte-Carlo simulations,
the system is quenched directly from a very high temperature (random initialization) to the desired temperature. The dynamics evolves for a certain number of sweeps (each sweep consists in $N_w$ proposed weight flips), and is 
finally sampled to evaluate thermodynamic quantities such as energy, spin glass order parameters. We run the dynamics in a total of $10^4$ sweeps, and after relaxation, $500$ configurations are collected for numerical
computations. The hyper-parameters of Monte-Carlo sampling are chosen to take the trade-off between accuracy and computational cost for a system of $\mathcal{O}(10^3)\sim\mathcal{O}(10^4)$ degrees of freedom.

We also perform simulated annealing experiments with $\tau$ Monte-Carlo sweeps per temperature. The temperature is decreased according to the set $\{2T_f,2T_f-\Delta T,\ldots,T_f\}$ where $\Delta T=0.1$.
We change the value of $\tau$ to study the temperature effects on the energy level that can be reached by the simulated annealing.

\begin{acknowledgments}
 We would like to thank Taro Toyoizumi for discussions. We also thank the referee for the stimulating comments. This research was supported by AMED under Grant Number JP18dm020700 and the start-up budget 74130-18831109 of
 the 100-talent-program of Sun Yat-sen University.
\end{acknowledgments}


\end{document}